\documentclass[letterpaper, 10 pt, conference]{ieeeconf}  

\IEEEoverridecommandlockouts                              

\usepackage{graphicx}
\usepackage{cite}
\usepackage{amsmath,amssymb,amsfonts}
\usepackage{algorithmic}
\usepackage{textcomp}
\usepackage{amsmath,amsfonts,amssymb}
\usepackage{algorithmic}
\usepackage{algorithm}
\usepackage{array}
\usepackage{bbm}
\usepackage[caption=false,font=footnotesize,labelfont=footnotesize,textfont=footnotesize]{subfig}
\usepackage{textcomp}
\usepackage{stfloats}
\usepackage{url}
\usepackage{verbatim}
\usepackage{graphicx}
\usepackage{booktabs}
\usepackage{threeparttable}
\usepackage{color}
\usepackage[table]{xcolor}
\usepackage{diagbox}
\usepackage{multirow}
\usepackage{cite}
\usepackage{flushend}
\usepackage{threeparttable}
\usepackage[hyphenbreaks]{breakurl}
\makeatletter
\let\NAT@parse\undefined
\makeatother
\usepackage{hyperref}
\hypersetup{pdfstartview=FitH,
            colorlinks=true,
            linkcolor=red,
            anchorcolor=blue,
            citecolor=green
            }
\newcounter{RNum}
\renewcommand{\theRNum}{\arabic{RNum}}
\newcommand{\Remark}{\noindent\textit{\textbf{Remark}~\refstepcounter{RNum}\textbf{\theRNum}}: }

\title{\LARGE \bf
Conditional Generative Denoiser for Nighttime UAV Tracking
}

\author{
 Yucheng Wang$^{1}$,
 Changhong Fu$^{2,*}$,
 Kunhan Lu$^{2}$,
 Liangliang Yao$^{2}$, and
 Haobo Zuo$^{3}$
\thanks{
*Corresponding author
}
\thanks{
$^{1}$Y. Wang is with the School of Electronic and Information Engineering, Tongji University, Shanghai 201804, China.
}%
\thanks{
$^{2}$C. Fu, K. Lu, and L. Yao are with the School of Mechanical Engineering, Tongji University, Shanghai 201804, China.
{\tt\small Email: changhongfu@tongji.edu.cn}
}%
\thanks{
$^{3}$H. Zuo is with the Department of Computer Science, University of Hong Kong, Hong Kong 999077, China.
}%
}

\begin{document}

\maketitle
\thispagestyle{empty}
\pagestyle{empty}

\begin{abstract}

State-of-the-art (SOTA) visual object tracking methods have significantly enhanced the autonomy of unmanned aerial vehicles (UAVs). However, in low-light conditions, the presence of irregular real noise from the environments severely degrades the performance of these SOTA methods. Moreover, existing SOTA denoising techniques often fail to meet the real-time processing requirements when deployed as plug-and-play denoisers for UAV tracking. To address this challenge, this work proposes a novel conditional generative denoiser (CGDenoiser), which breaks free from the limitations of traditional deterministic paradigms and generates the noise conditioning on the input, subsequently removing it. To better align the input dimensions and accelerate inference, a novel nested residual Transformer conditionalizer is developed. Furthermore, an innovative multi-kernel conditional refiner is designed to pertinently refine the denoised output. Extensive experiments show that CGDenoiser promotes the tracking precision of the SOTA tracker by 18.18\% on DarkTrack2021 whereas working 5.8 times faster than the second well-performed denoiser. Real-world tests with complex challenges also prove the effectiveness and practicality of CGDenoiser. Code, video demo and supplementary proof for CGDenoier are now available at:  \url{https://github.com/vision4robotics/CGDenoiser}.

\end{abstract}

\section{Introduction}

Vision-based unmanned aerial vehicle (UAV) tracking has become a popular topic owing to its widespread adoptions in robotics intelligence such as vision-based unmanned aerial manipulator approaching~\cite{Zheng2023ScaleAS}, intelligent offshore mooring~\cite{Ramos2022EKFBV}, aerial fueling~\cite{Gao2023DroguePM}, and so forth. 
However, problems arise in low-light scenes where intricate noise leads to poor tracking performance, especially aggravated after low-light enhancement~\cite{Ye2021DarkLighterLU,Ye2022TrackerMN}.
The increasing applications of vision-based UAVs demand trackers equipped with effective and practical real-noise denoisers.

\begin{figure}[t]
\centering
\includegraphics[width=1\linewidth]{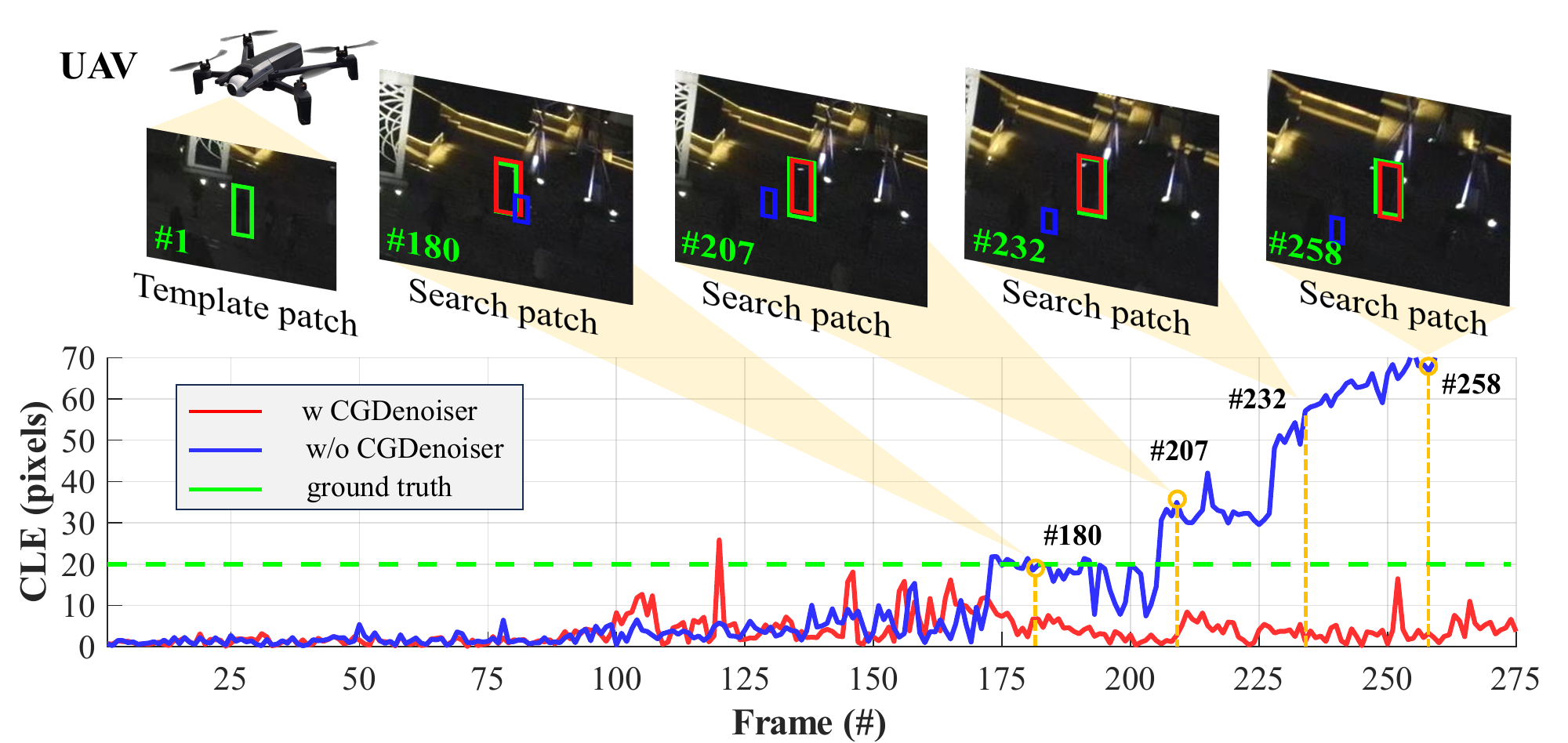}%
\vspace{-5pt}
\caption{
Tracking performance comparison in a typical dark scene with the proposed CGDenoiser utilized (in \textbf{\textcolor{red}{red}}) or not (in  \textbf{\textcolor{blue}{blue}}). The center location error (CLE) curves between predicted locations and ground truth bounding boxes are also exhibited. CGDenoiser raises tracking robustness in low-light conditions remarkably.
}
\label{fig:cle_comp}
\vspace{-18pt}
\end{figure}

To continuously access the location of an object, the state-of-the-art (SOTA) trackers are given their initial state as a reference and repeatedly estimate its location in the following frames by feature extraction and comparison~\cite{Kang2023ExploringLH,Chen2023HighPT}.
Recent years have witnessed a dramatic increase in the success rate as well as precision of tracking performance crediting persistent explorations in tracking methods.
Most early works~\cite{An2023LearningSR,Xing2022SiameseTP,Fu2024SAMDA} concentrate on improving tracking performance under well-illuminated circumstances. 
However, images captured under low-light circumstances suffer from intricate real noise and consequent information loss, which leads to poor generalization and severe performance decline of these SOTA trackers.
Several plug-and-play low-light enhancers~\cite{Ye2021DarkLighterLU,Ye2022TrackerMN} have been proposed and assist in brightening up the nighttime vision in UAV tracking. 
\textbf{\textit{Although SOTA trackers on UAVs equipped with low-light enhancers yield better performance on nighttime tracking, the noise is still intensified by enhancers and thus limits the tracking performance promotion. As such, it is essential to apply nighttime denoising techniques.}}

Nighttime denoising requires the model to process noisy frames captured under low-light circumstances and output clean images.
Traditional denoising methods concentrate on Gaussian denoising~\cite{Dabov2007ImageDB, Gu2014WeightedNN, Guo2019TowardCB}, training models to map images with Gaussian noise to corresponding noise-free images.
Nevertheless, the real noise with uncertainty does not strictly conform to the Gaussian distribution, which makes traditional denoisers generalize poorly in real-noise scenarios. Afterwards, the emergence of quality datasets~\cite{Abdelhamed2018AHQ,Plotz2017BenchmarkingDA} facilitates the development of deep learning-based denoising methods~\cite{Mehri2021MPRNetMP,Wang2022UformerAG,Zamir2022RestormerET}. Since the exact distribution of real noise is unknown, most methods design deep neural networks to acquire the end-to-end mapping from noisy images directly to clean images, compelling models to produce constant content for each specific image input.
However, the direct mapping retains considerably redundant information irrelevant to the noise and
the heavy network structure slows down processing speed, leaving trackers impractical in UAV tracking applications. 
Despite achieving higher scores on corresponding benchmarks, the practicality is limited in actual UAV tracking applications. \textbf{\textit{Therefore, there's a pressing need to design a high-performance and more practical denoiser for nighttime UAV tracking.}}

\begin{figure}[t]
\centering
\includegraphics[width=0.9\linewidth]{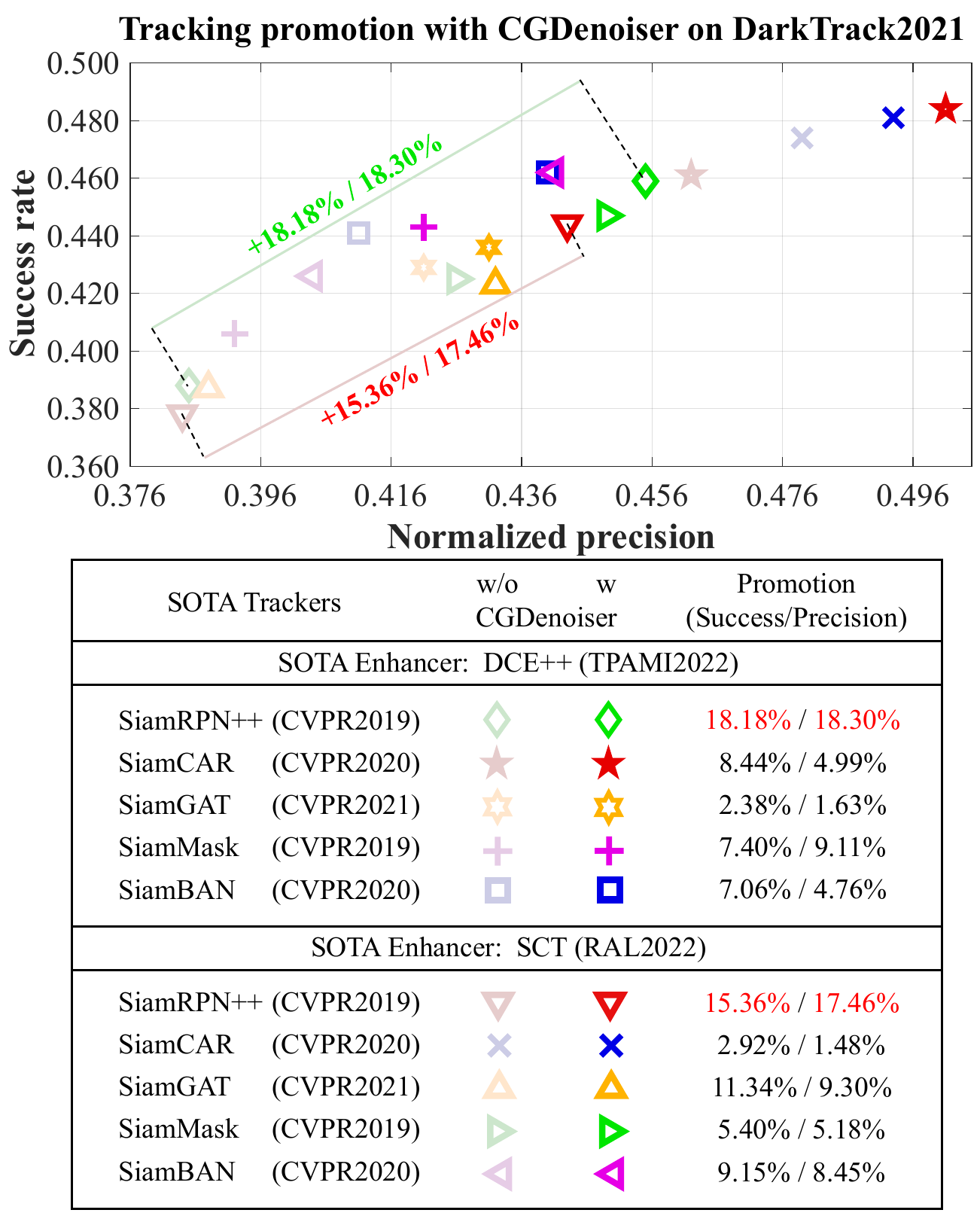}%
\caption{Overall performance promotion on the DarkTrack2021 benchmark~\cite{Ye2022TrackerMN} for leading-edge trackers~\cite{Li2019SiamRPN++EO,Guo2021GraphAT,Chen2020SiameseBA,Guo2020SiamCARSF,Wang2019FastOO} enhanced by different SOTA enhancers~\cite{Li2022LearningTE,Ye2022TrackerMN}.
Symbols with the \textbf{dark} color represent original enhanced trackers, while the \textbf{light} ones represent enhanced trackers with the denoising support of CGDenoiser, which significantly improves tracking success rate and precision.
}
\vspace{-15pt}
\label{fig:effectiveness}
\end{figure}

Supervised training is utilized in the conditional generative model (CGM) to form structured outputs~\cite{Sohn2015LearningSO}. Unlike non-generative methods, constraints from supervised training loss are loosened due to sample randomness, guiding CGM to apprehend the distribution of outputs from a macro perspective. Inspired by this, CGM is introduced in the proposed conditional generative denoiser (CGDenoiser) to learn the intractable distribution of real noise, devising a more practical plug-and-play denoiser for further nighttime UAV tracking. Guided by supervised training, CGDenoiser is capable of interpreting and conditionalizing the context of each input then correspondingly generating its signal-dependent noise to remove.
Generative design alleviates the limitation of pixel-wise loss, which frees CGDenoiser from redundant content limitation, drawing attention back to the original distribution of real noise.
Further, the residual structure allows CGDenoiser to directly inspect the formation of noise instead of focusing on content details. 
Additionally, to address the issue of noise amplification caused by unreasonable post-processing of practical cameras, a multi-kernel conditional refiner (MKCR) post-processing method is designed.
It specifically generates post-processing convolutional kernels for each input image to refine the preliminarily denoised output from previous steps and obtain the final result.
In addition, a novel nested residual Transformer conditionalizer (NRTC) is devised to extract conditional representation from input images as well as to accelerate processing speed by downsampling.
As shown in Fig.~\ref{fig:effectiveness}, CGDenoiser dramatically improves the performance of multiple SOTA trackers equipped with low-light enhancers.

The contributions of this work lie four-fold:
\begin{enumerate}
\item{A real-time and real-noise denoiser is proposed by creatively generating the noise from the input, more practically removing real-noise in nighttime UAV vision and significantly improve UAV tracking performance.}
\item{An innovative NRTC is introduced to better align the dimension of inputs and accelerate inference by producing more representative condition maps for generation}
\item{A novel MKCR is invented to flexibly and adaptively generate signal-dependent kernels for convolutional refinement in the post-processing step.}
\item{Extensive experiments demonstrate that CGDenoiser can boost nighttime UAV tracking capability by adaptively estimating and removing the real noise in the nighttime vision of UAV trackers with enhancers whereas keeping high processing speed onboard. Real-world tests serve to confirm its practicality and efficacy.}
\end{enumerate}

\section{Related Works}

\label{sec:Related Works}
\subsection{Vision-based UAV Tracking}
\label{sec:UAV Tracking}

Most current SOTA methods for object tracking follow the Siamese paradigm, which can be typically classified into two categories according to whether they use pre-defined anchors or not. The former ones~\cite{Li2019SiamRPN++EO,Li2018HighPV} utilize region proposal networks to predict bounding boxes from pre-defined anchors, whereas the latter ones~\cite{Chen2020SiameseBA,Guo2021GraphAT,Guo2020SiamCARSF} devise anchor-free pipelines to avoid the uncertainties incurred by hyperparameters. Further researches aggregate the upsides of both categories and show high efficiency in UAV tracking based on more quality and precise anchors.
Most of these methods are designed for well-lit scenes. However, when generalizing on low-light images, the performance loss is significant, resulting in a drastic decrease in tracking results. To address illumination issues, some plug-and-play low-light enhancers~\cite{Ye2021DarkLighterLU,Ye2022TrackerMN} have been proposed and applied. Whereas the use of enhancers does improve nighttime tracking performance, the noise is also amplified by the enhancer, which in turn limits the promotion.

\subsection{Image Denoising}\label{sec:Image Denoising}

The development of supervised image denoising method is greatly influenced by the proposal of new datasets~\cite{Abdelhamed2018AHQ,Plotz2017BenchmarkingDA}. Traditional image denoising methods are mostly based on the assumption of Gaussian distribution of image noise. Methods like BM3D~\cite{Dabov2007ImageDB} and WNNM~\cite{Gu2014WeightedNN} stand out in non-learning-based algorithms. With the development of deep learning, more researchers were engaged in devising models using convolutional neural networks (CNN) for denoising~\cite{Guo2019TowardCB}, which proved to be more effective than most non-learning-based methods. 
Despite gaining more precise predictions on the patches of different noise level, such methods still cope with Gaussian noise due to the shortage of high-quality real denoising datasets. Since new real-noise training datasets were published~\cite {Abdelhamed2018AHQ,Plotz2017BenchmarkingDA}, focus has been transferred to the real-denoising task. Deep learning-based models, \emph{e.g.}, Restormer~\cite{Zamir2022RestormerET}, MPRNet~\cite{Mehri2021MPRNetMP}, and Uformer~\cite{Wang2022UformerAG}, deepening and widening the networks do reach considerably high metrics scores on corresponding benchmarks. Compared to traditional Gaussian denoising, real-world denoising methods generalize much better in practical application, simply because the distribution of noise in the real world differs significantly from the Gaussian distribution. However, high score does not mean better application in the real world. Problems like unsatisfying tracking precision and slow processing speed narrow the application. To better integrate low-light enhancers, trackers, and denoisers into nighttime UAV operations, it is necessary to design a more deployable and practical plug-and-play denoiser.

\section{Methodology}

\begin{figure*}[ht]
\vspace{-5pt}
\centering
\includegraphics[width=1\linewidth]{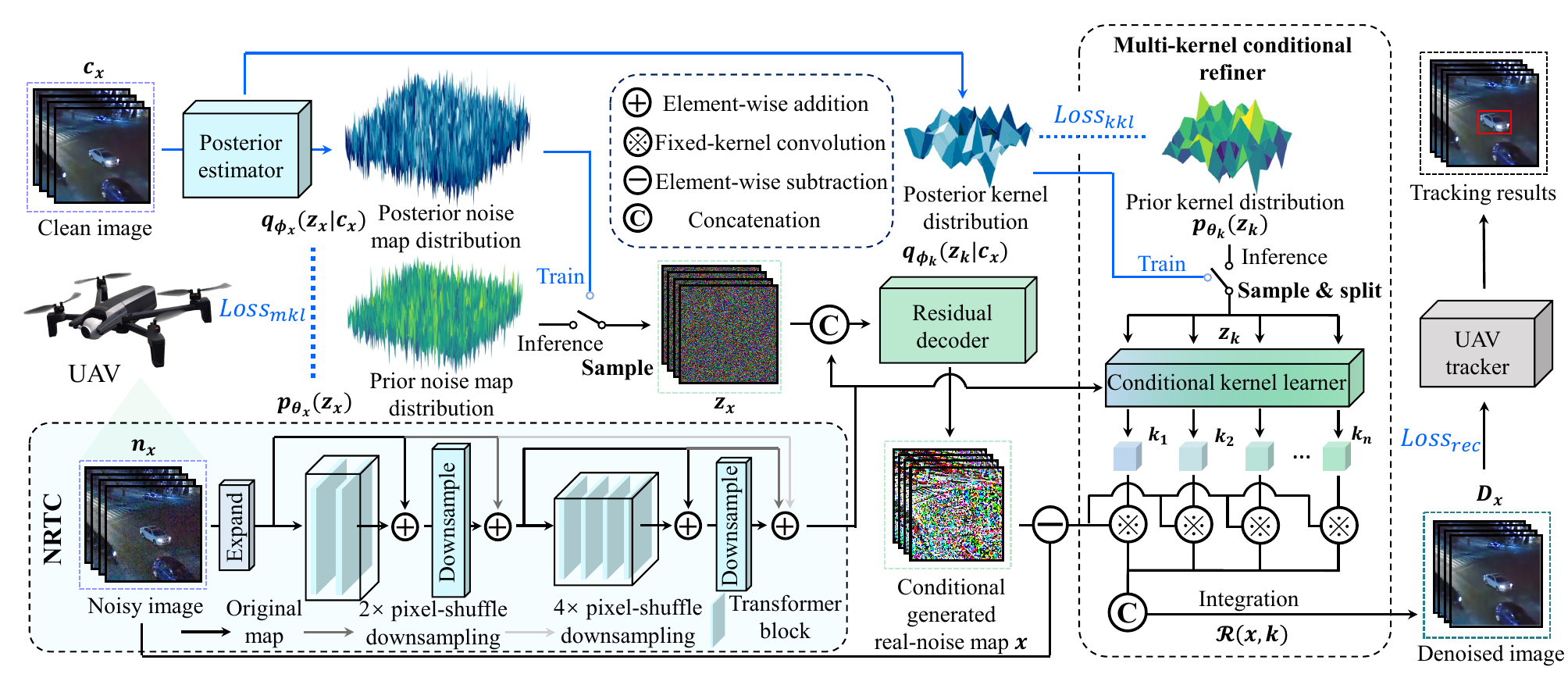}
\vspace{-25pt}
\caption{The pipeline of the proposed conditional real-noise generative denoiser. 
Given a low-light noisy image patch, CGDenoiser dedicates to conditionally generate its noise map then remove it, resulting in dramatic improvement of UAV trackers performance.
}
\label{fig:main}
\vspace{-14pt}
\end{figure*}

The pipeline of the proposed method is demonstrated in Fig.~\ref{fig:main}. Noisy frames are compressed into condition maps through NRTC.
In the training stage, clean images serve for the estimation of posterior distribution of noise map and kernels. The posterior sample noise are concatenated with the condition map from NRTC and decoded into real-noise maps. The sample kernels, together with the condition map are fed into conditional kernel learner to produce conditional kernels for further refinement. The posterior and prior distributions are drawn closed during training. 
After noise removal, the preliminarily denoised patches are refined by conditional generated kernels in MKCR.
In the inference stage, both posterior distributions are replaced by the prior ones, and others remain the same. The refined frames are finally sent for enhancement and tracking. Supplementary proof of the proposed method is now available at : \url{https://github.com/vision4robotics/CGDenoiser}.

\subsection{Conditional Dual Branch Generator}\label{sec:Conditional real-noise and kernel generator}

Denote the distribution of real noise as $q(x)$. Given noisy input $n_{x}$ as conditional prompt, the goal is to maximize the likelihood of conditional distribution $p_{\theta}(x|n_{x})$.
K. Sohn \emph{et al}.~\cite{Sohn2015LearningSO} have demonstrated the variational lower bound of the log-likelihood as: 
\begin{equation}
\begin{split}
\log{p_\theta(x|n_x)}&\ge\mathbb{E}_{q_\phi(z|n_x,x)}[\log{p_\theta(x|n_x,z)}] \\
&-KL[q_\phi(z|n_x,x)||p_\theta (z|n_x)] \quad, 
\end{split}
\end{equation}
where $n_x$ and $z$ denote the noisy input and the corresponding latent variable respectively.

The pipeline in CGDenoiser is required to generate two components conditioned on each input frame: (i) signal-dependent real noise map, (ii) kernels for conditional refinement. 
The estimation of kernels $k$ for conditional refinement is similar to real-noise map $x$, whose reconstruction expectation can be written as:
\begin{equation}
\begin{split}
\nonumber
&\mathbb{E}_{q_{\phi_k}(z|n_x,k)}[\log{p_\theta(k|n_x,z)}] \quad.
\end{split}
\end{equation}

Given real-noise map $x$ and conditional kernels $k$, the restored image $D_x$ is defined as:
\begin{equation}
\begin{split}
&D_x = \mathcal{R}(n_x-x,k) \quad, 
\end{split}
\end{equation}
where $\mathcal{R}$ denotes the fixed-kernel convolutional operation and integration in MKCR.

During the training stage, the clean image is considered to contain sufficient prior knowledge for the posterior latent distribution $q_{\phi}$. To compress abundant prior as well as simplify the training pipeline, the posterior distribution of latent variable $q_{\phi_x}(z_x|n_x, x)$ and $q_{\phi_k}(z_K|n_x, k)$ is implemented as $q_{\phi}(z|c_x)$.  Therefore the reconstruction terms are rewritten below:
\begin{equation}
\begin{split}
\nonumber
&\mathbb{E}_{q_{\phi_x}(z_x|c_x)}[\log{p_\theta(x|n_x,z_x)}] \quad, \\
&\mathbb{E}_{q_{\phi_k}(z_k|c_x)}[\log{p_\theta(k|n_x,z_k)}] \quad.
\end{split}
\end{equation}

Taking the proposed conditionalizer into account, the final optimization objective is represented as follows:
\begin{equation}
\begin{split}
\label{equ:loss}
&\mathcal{L}(\phi_x,\phi_k,\theta)= \\
&-KL[q_{\phi_x,\phi_k}(z_x,z_k|c_x)||p_\theta (z_x,z_k|\mathrm{NRTC}(n_x))] \\
&+\mathbb{E}_{q_{\phi_x,\phi_k}(z_x,z_k|c_x)}[\log{p_\theta(D_x| \mathrm{NRTC}(n_x),z_x,z_k)}] \quad.
\end{split}
\end{equation}

The first term in Eq. (\ref{equ:loss}) functions as narrowing the KL-divergence between the prior latent distribution $p_\theta (z_x,z_k|\mathrm{NRTC}(n_x))$ and the posterior latent distribution $q_{\phi_x,\phi_k}(z_x,z_k|c_x)$, corresponding to $\mathit{Loss_{mkl}}$ and $\mathit{Loss_{kkl}}$ in Fig.~\ref{fig:main}. Assuming $z_x$ and $z_k$ follow independent multivariate Gaussian distribution, the KL-divergence could be explicitly expressed and narrowed as in classical variational autoencoder (VAE)~\cite{kingma2013AutoEV}. The latter term, associated with $\mathit{Loss_{rec}}$ in Fig.~\ref{fig:main}, is interpreted as reconstruction loss, which can be quantified and optimized by Monte Carlo simulation~\cite{Sohn2015LearningSO} and pixel-wise loss.

\begin{figure}[!t]
\includegraphics[width=1\linewidth]{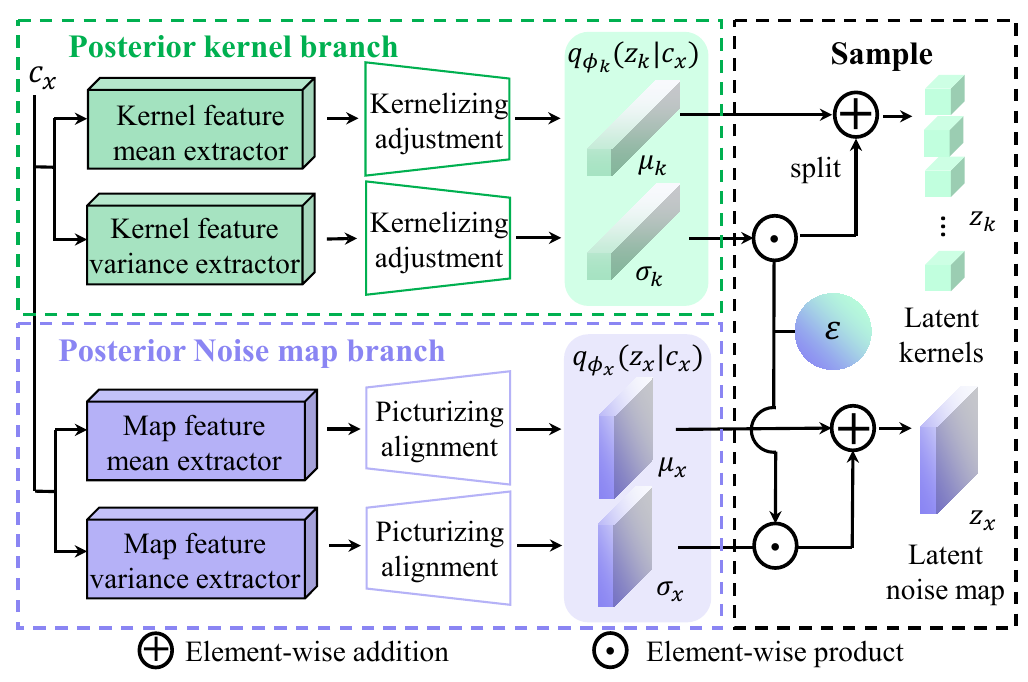}
\vspace{-20pt}
\caption{Illustration of the dual-branch posterior estimator (DPE) in the training stage. The input $c_{x}$ denotes the clean image and the output $z_{k}$ and $z_{x}$ represent posterior latent variables for kernels $k$ and real-noise map $x$ respectively. DPE consists of two posterior estimation branches, each of which encodes a clean image into the latent distribution of either kernels or noise map by feature extraction and alignment.
}
\label{fig:PE}
\vspace{-10pt}
\end{figure}

Based on the above deduction, a dual-branch posterior estimator (DPE) is designed to assist model training. As shown in Fig.~\ref{fig:PE}, the DPE encoder decomposes the estimation of posterior latent distribution into two branches, each of which is further split in two, mapping the mean and variance of Gaussian distribution respectively through feature extraction and dimension alignment. Kernels for post-processing refinement lay emphasis on the global image attributes whereas the preliminarily generated noise map attaches more importance to signal details. Therefore, the segmentation of posterior estimation enables CGDenoiser to more attentively extract features from different perspectives. To make the network differentiable, reparameterization is used in the sample stage as illustrated on the right of Fig.~\ref{fig:PE}.

\Remark Due to the randomness in the sampling process of CGDenoiser, constraints from supervised training are conspicuously loosened. The network is instead guided to explore further insight into the real noise distribution from a more macro perspective, which effectively reduces model's dependence on the training dataset and frees the network from overfitting.

\subsection{Multi-kernel Conditional Refiner}\label{sec:Multi-kernel Conditional Refiner}

To further improve the quality of the preliminarily denoised results from residual subtraction, an MKCR is designed for post-processing. As shown on the right of Fig.~\ref{fig:main}, the conditional kernel learner adaptively produces kernel stack $\mathbf{k}\in \mathbb{R}^{(3\times num_k)\times size_k\times size_k}$ conditioned on the processed noisy input $\mathrm{NRTC}(n_x)$, which then serve as kernels for fixed-kernel convolution on the former denoised results. Given preliminarily denoised image $d_x = n_x - x, \ d_x\in \mathbb{R}^{3\times H\times W}$ and conditional generated kernels $\mathit{k_i}\in \mathbb{R}^{3\times size_k\times size_k}$ split from $\mathbf{k}$ and denoting fixed-kernel convolution operation as $\bf{\ast} \ $, MKCR can be formulated as:
\begin{equation}
\begin{split}
D_x &=\mathcal{R}(d_x,\mathbf{k}) \quad \\
&=\mathcal{I}(d_x \ast k_1, d_x \ast k_2, ..., \ d_x \ast k_{num_k}) \quad,
\end{split}
\end{equation}
where $\mathcal{I}$ represents the integration of convolution maps. $num_k$ and $size_k$ are respectively the number and the size of kernels. 

\Remark Equipped with MKCR for post-processing, CGDenoiser manages to refine denoised outputs more flexibly and adaptively way conditioned on input signals, conquering volatile problems like overexposure, and color distortion due to inappropriate general post-processing in cameras.

\subsection{Nested Residual Transformer Conditionalizer}\label{sec:Nested Residual Transformer Conditionalizer}

Due to the fact that the latent variables $z$ are sampled from a high-dimensional space whereas the noisy inputs $n_x$ stay low-level, misalignment occurs when combining $z$ and $n_x$ abruptly, leading to poor decoding results. Besides, $n_x$ of original resolution disastrously slows down the processing speed, leaving denoiser impractical as a UAV plug-and-play component. To solve both problems, an NRTC is devised to align latent variable $z$ and condition $n_x$ as well as lighten the whole structure by downsampling operation. 
As shown in Fig.~\ref{fig:main}, NRTC consists of a channel expansion layer, conditional extractors, and downsampling blocks. Given input tensor $n_x$, the NRTC process is defined as:
\begin{equation}
\begin{aligned}
C^{0} &= \mathrm{Ep}(n_x)\quad, \\
C^{1} &= \mathrm{DS}(\mathrm{CE}(C^{0})+C^{0})+\mathcal{P}^{2\times}(C^{0})\quad, \\
C^{2} &= \mathrm{DS}(\mathrm{CE}(C^{1})+C^{1})+\mathcal{P}^{2\times}(C^{1})\quad, \\
\mathrm{NRTC}(n_x) &= C^{2}+\mathcal{P}^{4\times}(C^{0})\quad,
\end{aligned}
\end{equation}
where $\mathrm{Ep}(\cdot):\mathbb{R}^{3\times H\times W}\to \mathbb{R}^{C\times H\times W}$ represents channel expansion; $\mathrm{DS}(\cdot)$ denotes downsampling layer; $\mathcal{P}^{\alpha\times}(\cdot)$ means pixel-unshuffle downsample operation with downscale factor $\alpha$; and $\mathrm{CE}(\cdot)$ denotes conditional extractor composed of basic Transformer blocks~\cite{Zamir2022RestormerET}.

\Remark The nested design enables NRTC to fuse features from multiple scales and flexibly handle information of different frequencies, filtering out redundant contents and condensing conditional representation, which is conducive to the combined interpretation of latent variable $z$ and condition $n_x$, and significantly increases processing speed.

\section{Experiments}

\begin{figure*}[!t]
\subfloat{\includegraphics[width=0.33\linewidth]{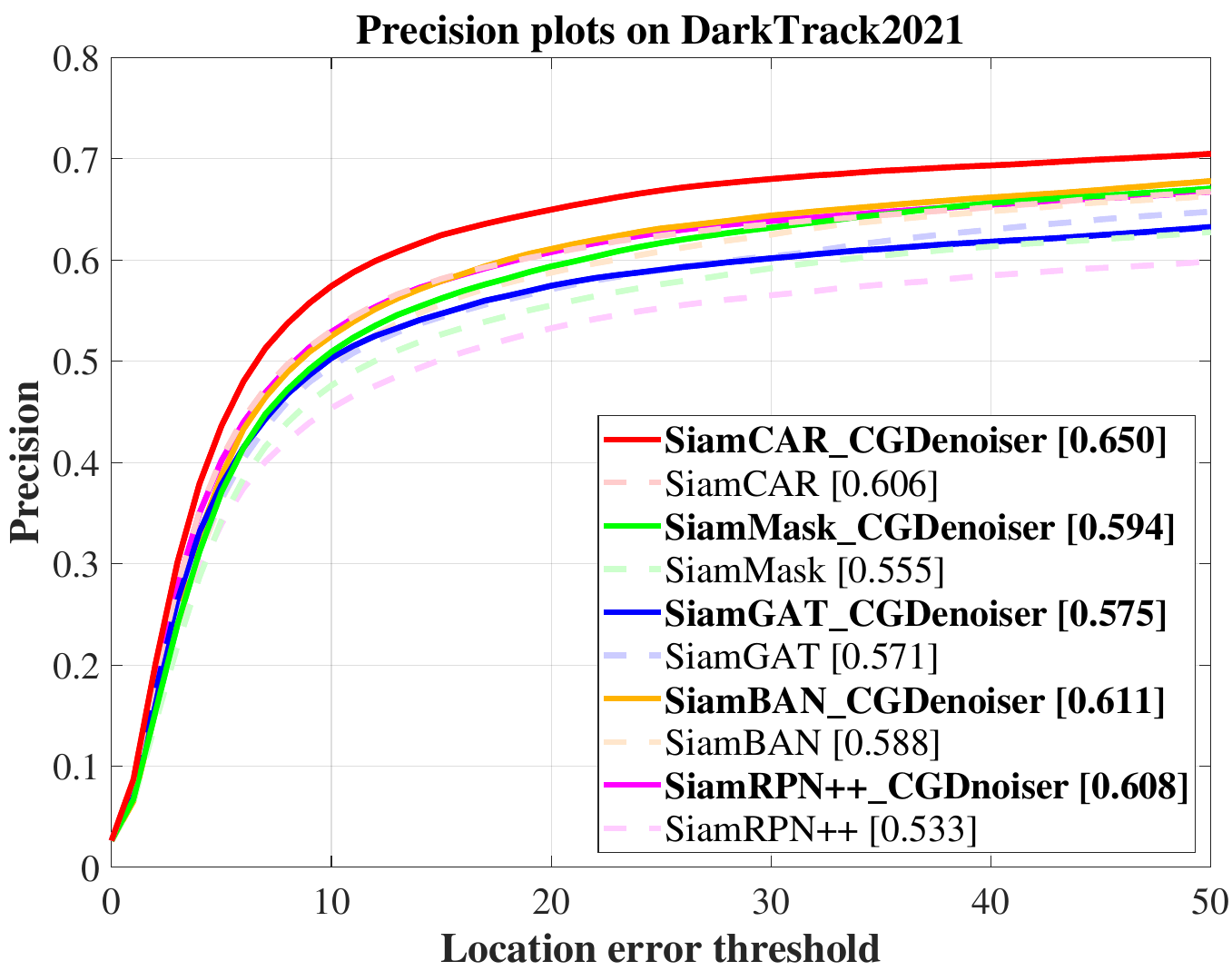}%
}
\hfil
\subfloat{\includegraphics[width=0.33\linewidth]{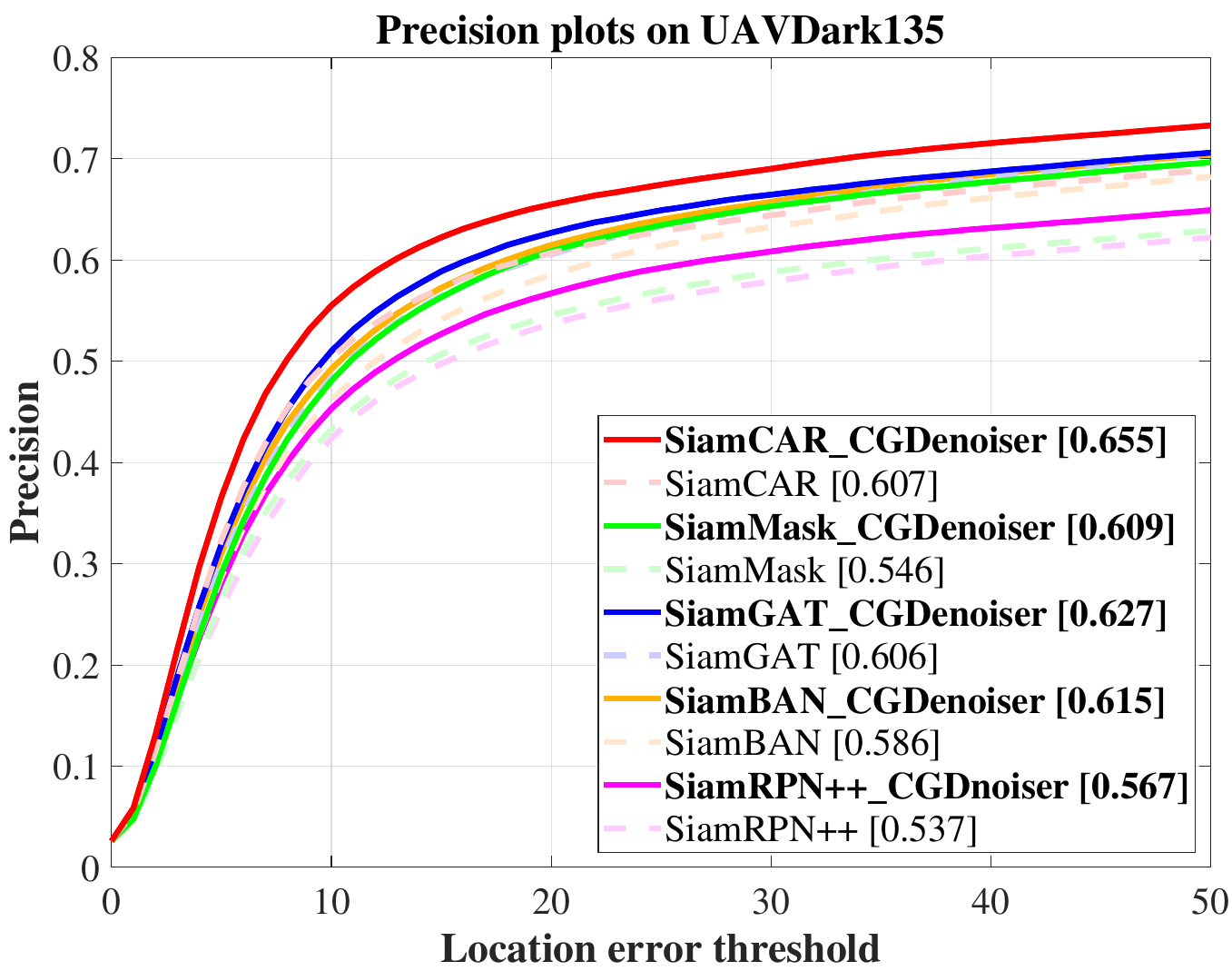}%
}
\hfil
\subfloat{\includegraphics[width=0.33\linewidth]{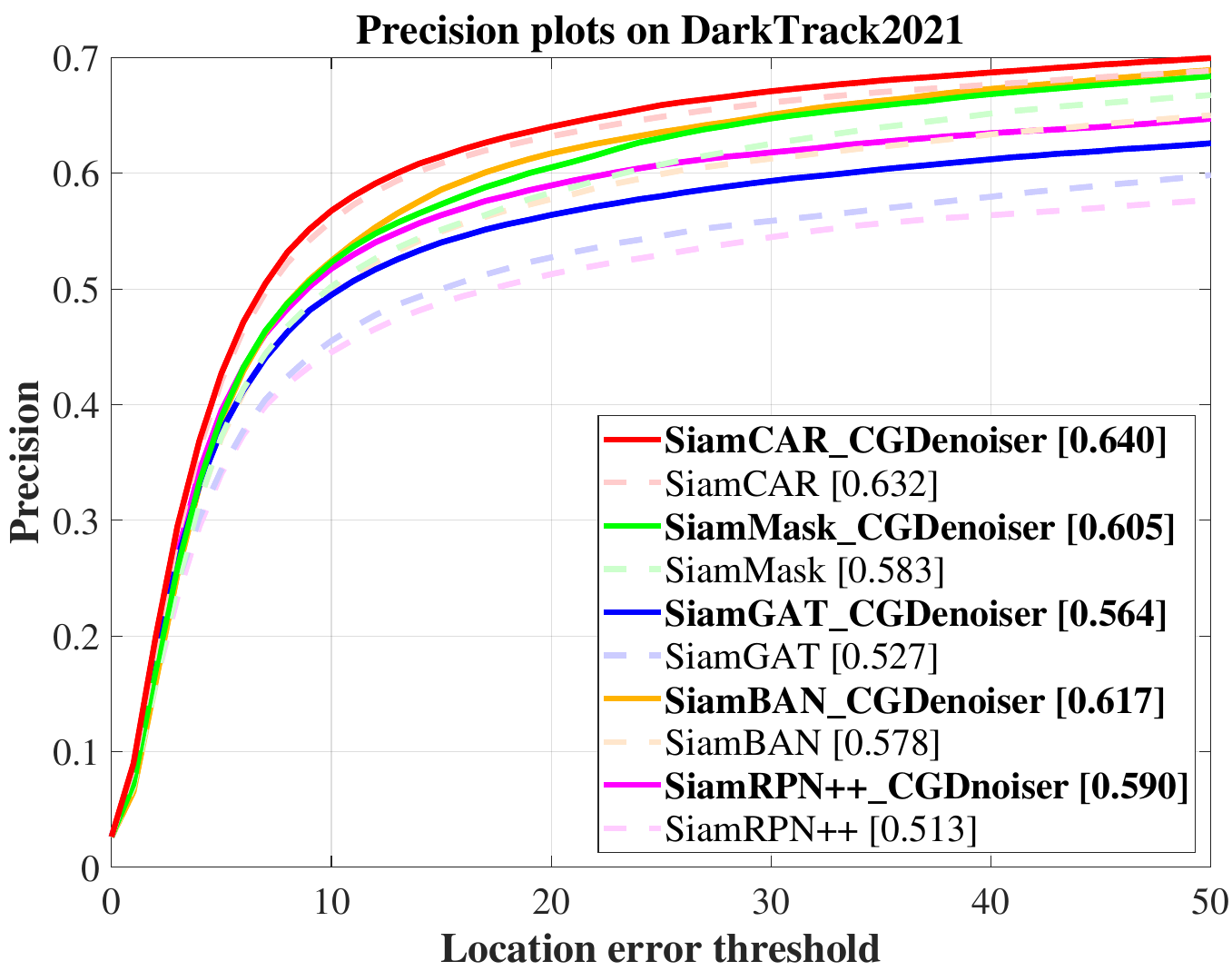}%
}
\hfil
\subfloat{\includegraphics[width=0.33\linewidth]{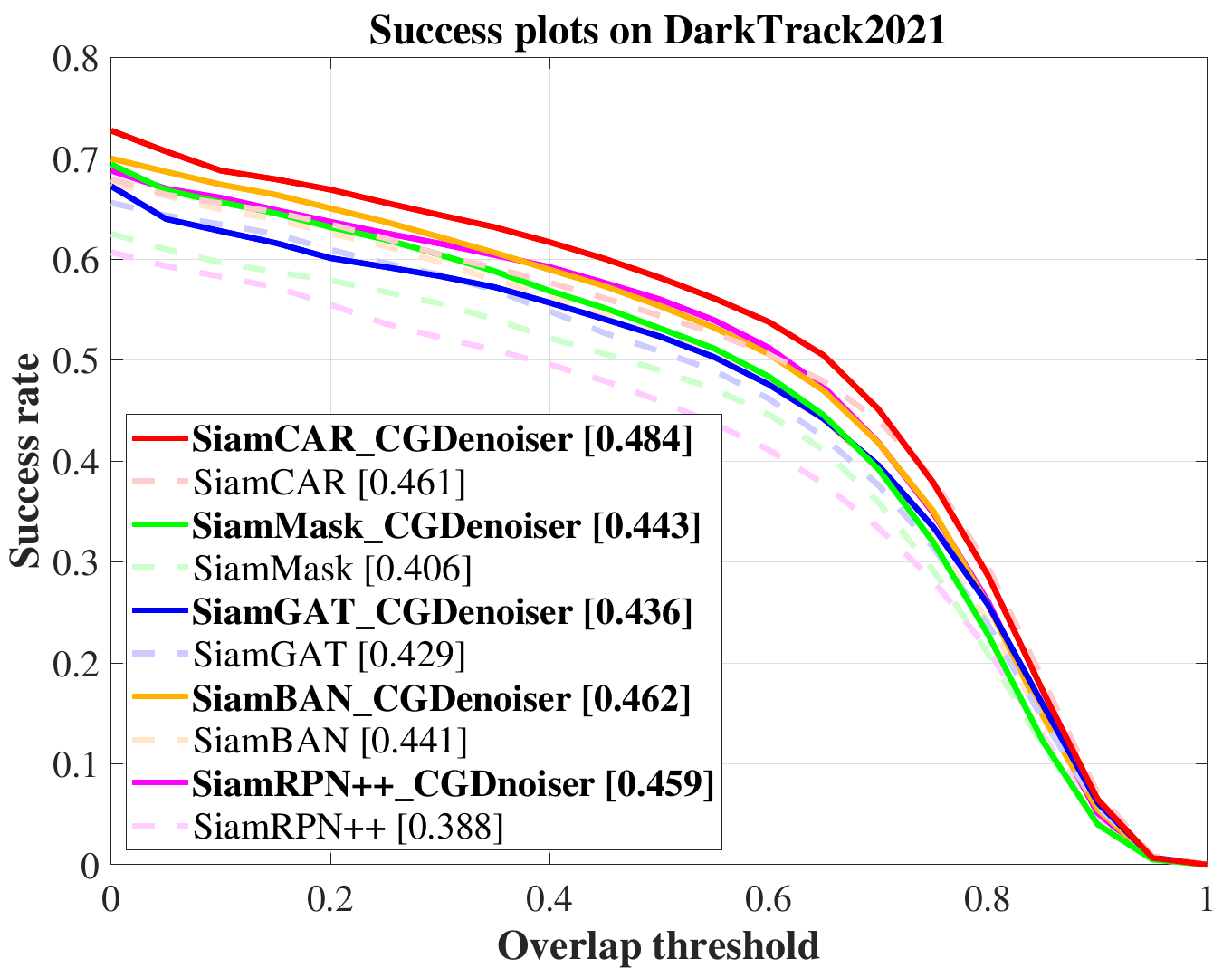}%
}
\hfil
\subfloat{\includegraphics[width=0.33\linewidth]{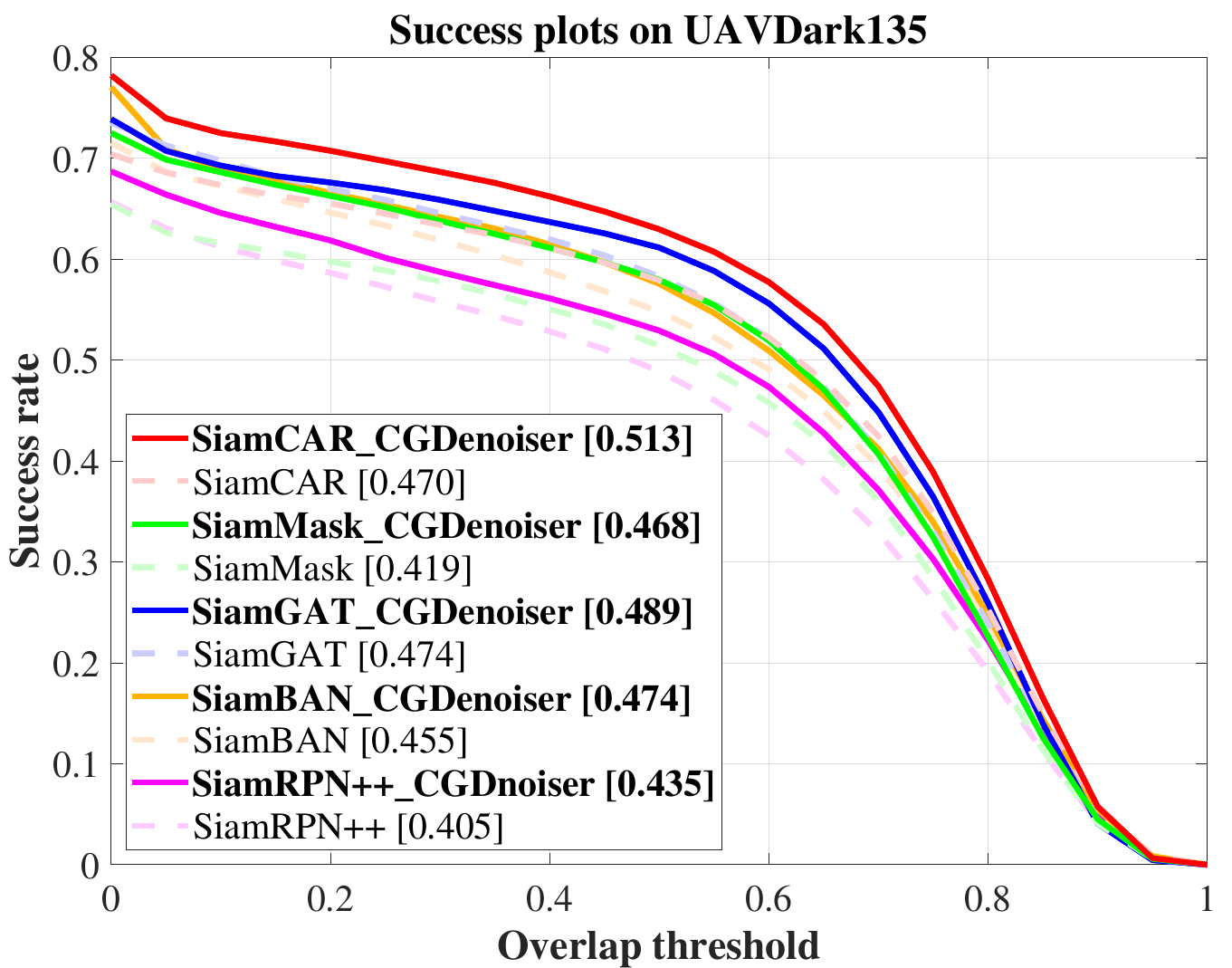}%
}
\hfil
\subfloat{\includegraphics[width=0.33\linewidth]{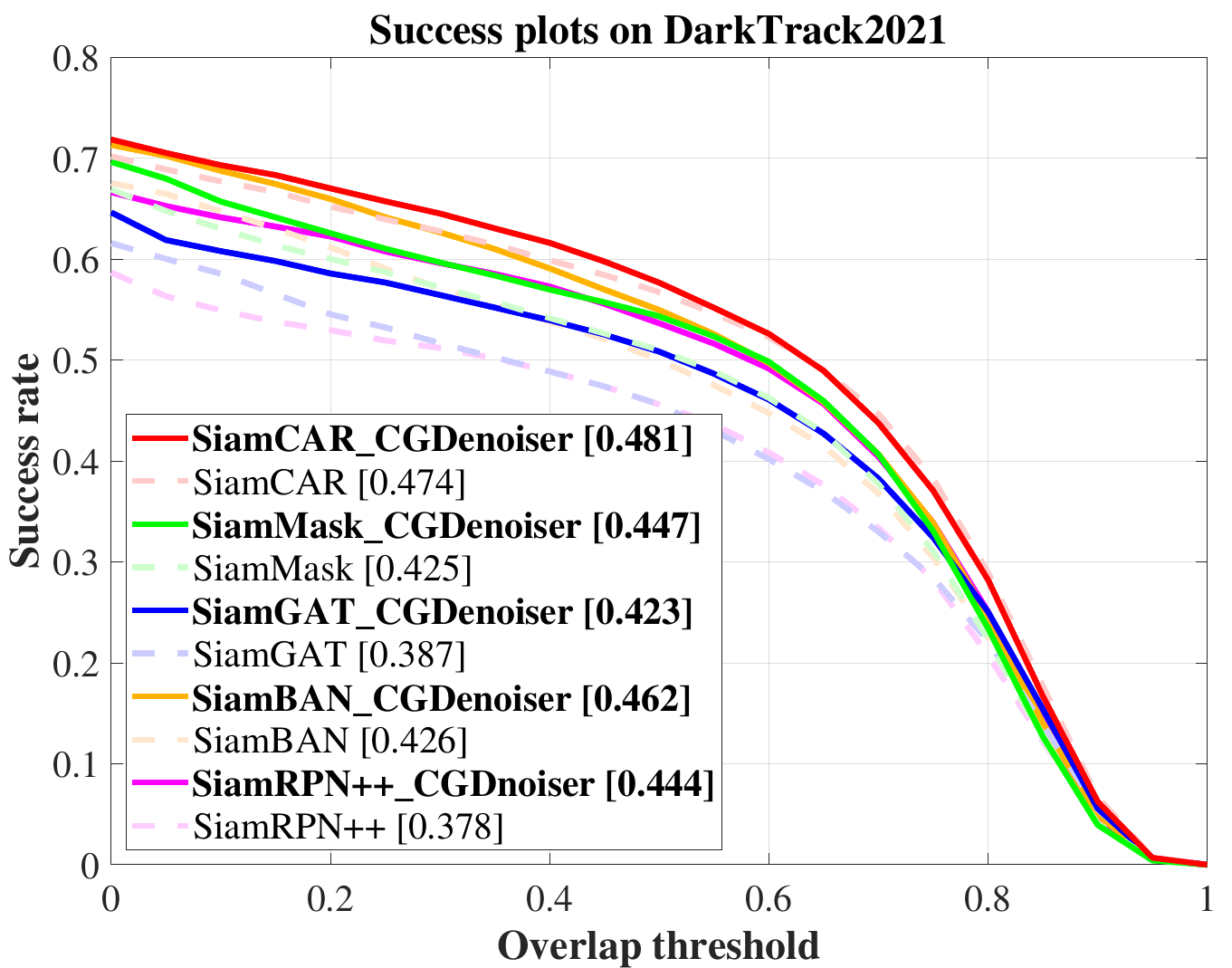}%
}
\hfil

\vspace{-6pt}
\caption{Tracking promotion of trackers on two authoritative benchmarks~\cite{Li2022Alldo,Ye2022TrackerMN}. To further demonstrate the effectiveness of CGDenoiser, trackers with different enhancers are evaluated (the first and second column of results are tested with DCE++~\cite{Li2022LearningTE}, while the last column with SCT~\cite{Ye2022TrackerMN}). Solid and dashed lines respectively represent the performance with and without CGDenoiser. Results with higher metrics are bolded.
}
\label{fig:rainbow}
\vspace{-15pt}
\end{figure*}

\subsection{Implementation Details}\label{Implementation Details}

CGDenoiser is trained with the assistance of DPE as the blue pipelines shown in Fig.~\ref{fig:main}. DPE narrows the distance between the posterior latent distribution $p_\theta (z_x,z_k|\mathrm{NRTC}(n_x))$ and the prior latent distribution $q_{\phi_x,\phi_k}(z_x,z_k)$ so that the residual decoder can produce reasonable outputs when inferencing without DPE.   
The kernel size and number of MKCR in CGDenoiser are $size_k=3$ and $num_k=20$ respectively.
The paired clean/noisy images for training are cropped from the SIDD dataset~\cite{Abdelhamed2018AHQ}.
A progressive training strategy is adopted to train CGDenoiser . AdamW optimizer ($\beta_{1}=0.9$, $\beta_{2}=0.999$) with weight decay $1e^{-4}$ and cosine annealing learning rate scheduler (from $3e^{-4}$ to $1e^{-6}$) are utilized for 600K iterations training.

Before inferencing in trackers, the inputs,  including the $287 \times 287$ search patch of each frame and the $127 \times 127$ template of the initial frame, are first enhanced by low-light enhancer~\cite{Li2022LearningTE,Ye2022TrackerMN} then denoised by CGDenoiser.

Experiments are conducted on a PC with 2 Tesla V100 GPUs and 96 Intel Xeon Platinum 8163 CPUs.

\subsection{Evaluation Metrics}\label{Evaluation Metrics}

The proposed plug-and-play denoiser serves for nighttime UAV tracking. To quantify the quality of tracking results, success rate and precision plots from one-pass evaluation (OPE)~\cite{Wu2013OnlineOT} are adopted for tracking performance evaluation. 
Specifically, success rate is gauged by the intersection over union (IoU) between the estimated bounding box and the ground truth bounding box. The success plot (SP) showcases the percentage of frames with an IoU exceeding a predefined maximum threshold. Typically, the area under the curve (AUC) on the SP is employed to rank tracker success rates.
Precision, on the other hand is quantified by the center location error (CLE) between the tracking output and the ground truth. We present the precision plot (PP) as the percentage of frames with a CLE below a specified threshold, with 20 pixels commonly used to evaluate tracker performance.
Besides tracking performance, denoising performance and processing speed are also evaluated. 
The peak signal-to-noise ratio (PSNR) of the test data from SIDD are adopted as the evaluation metric of denoising performance.

\subsection{Combined Evaluation of CGDenoiser with Trackers and Low-Light Enhancer}\label{Combination of CDT with Trackers and Low-Light Enhancers}

To demonstrate the effectiveness of CGDenoiser, combined tests with multiple SOTA UAV trackers enhanced by nighttime enhancers are conducted on two authoritative low-light tracking benchmarks, \emph{i.e.}, UAVDark135 and DarkTrack2021. As shown in Fig.~\ref{fig:rainbow}, the tracking precision and success rate of SOTA trackers both gain dramatic increase. Conspicuously, with the synergy of CGDenoiser, the tracking precision and success rate of SiamRPN++ enhanced by DCE++ on DarkTrack2021 mounts by \textbf{18.18\%} and \textbf{18.30\%} respectively. A similar remarkable boost happens on the combination of CGDenoiser and SiamMask enhanced by DCE++ on UAVDark135 (\textbf{11.38\%} / \textbf{11.69\%}) and SiamRPN++ enhanced by SCT on DarkTrack2021 (\textbf{15.36\%} / \textbf{17.46\%}). 

\begin{figure}[!t]
\centering
\includegraphics[width=1\linewidth]{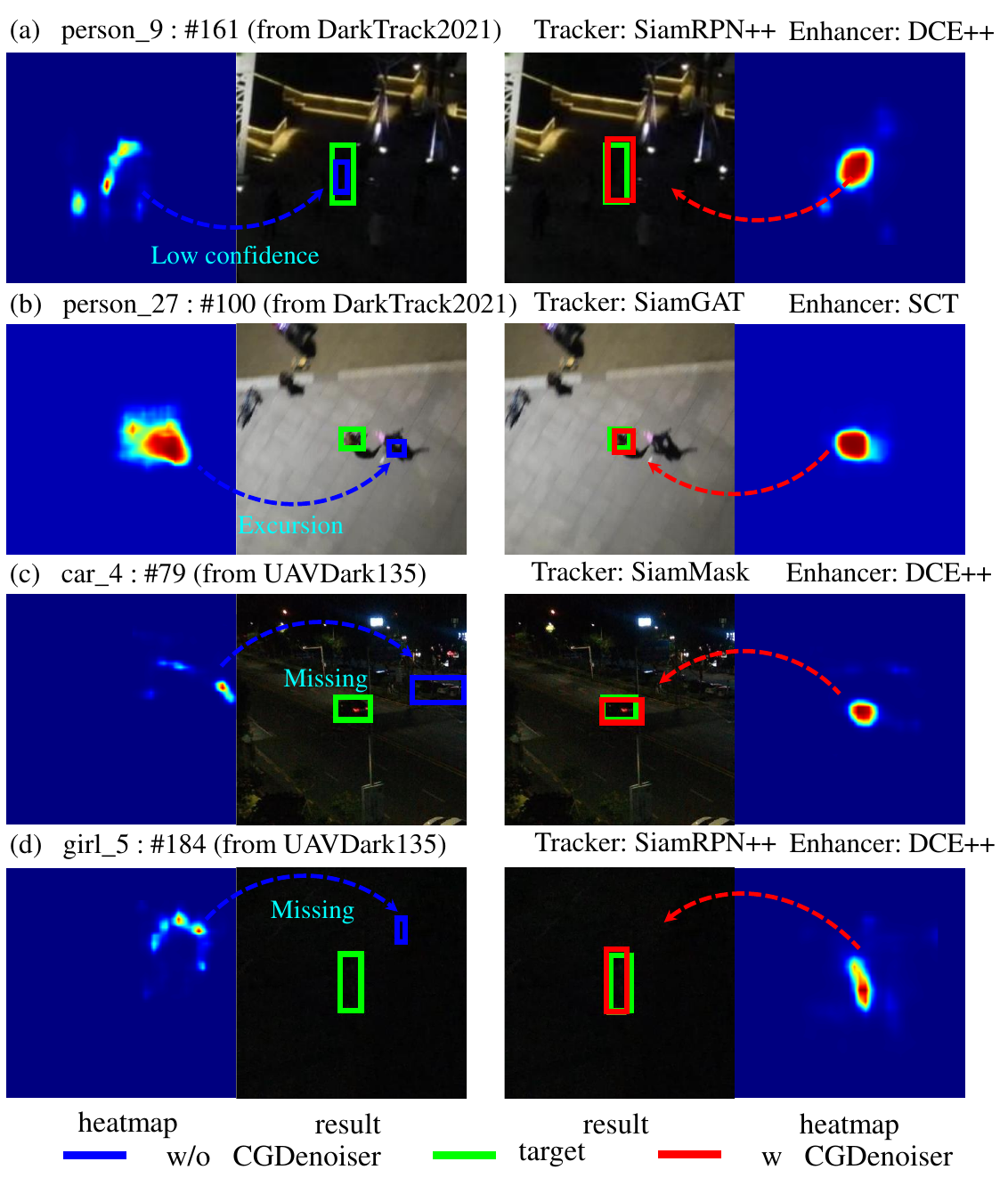}
\vspace{-20pt}
\caption{Visualization of the proposal region heatmap from the classification score of tracker backbone. The utilization of the proposed method effectively draws trackers' attention back to targets thus producing more precise predicting results.
}
\label{fig:heatmap}
\end{figure}

\begin{figure}[t]
\centering
\subfloat{\includegraphics[width=1\linewidth]{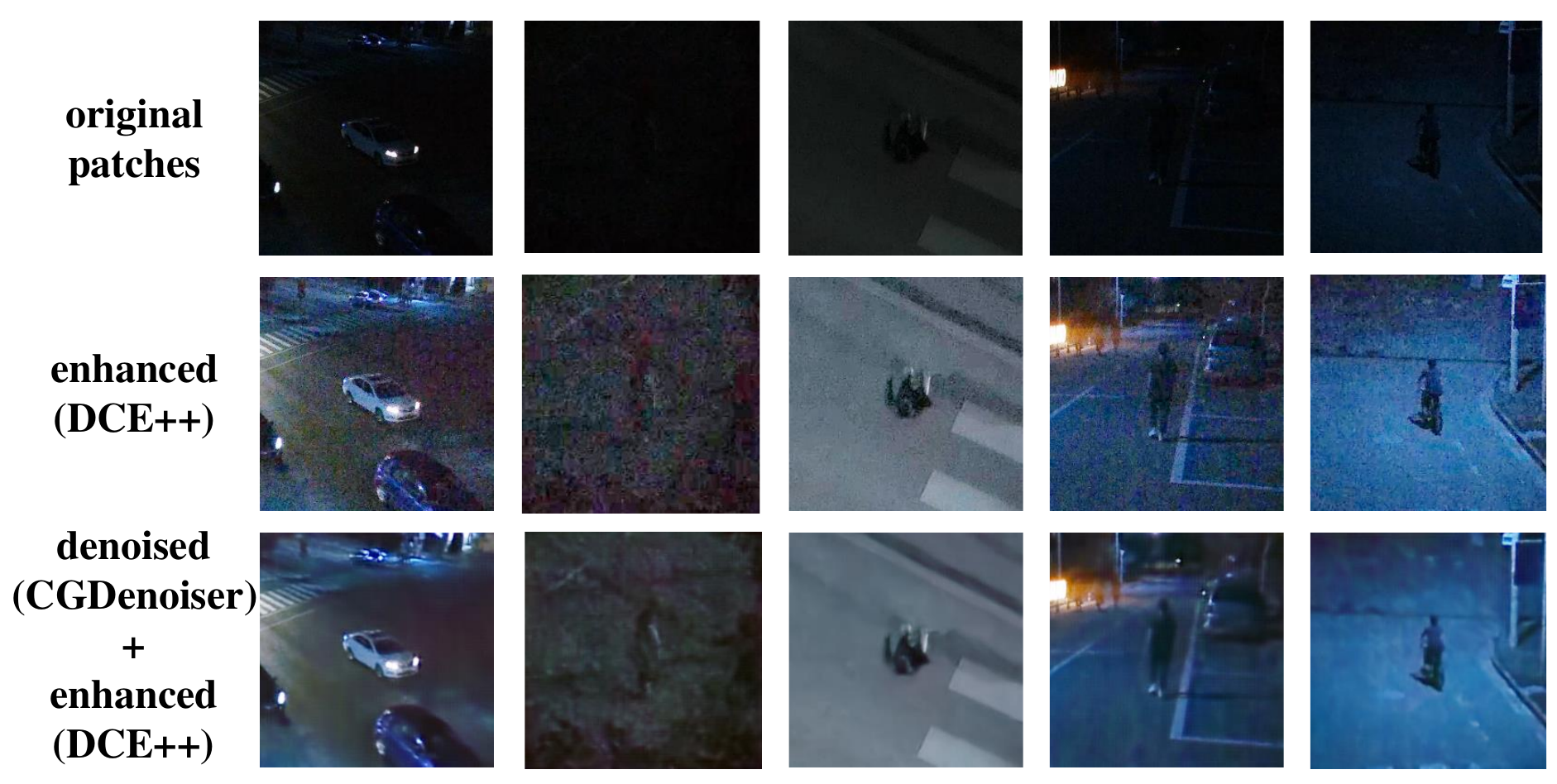}}%
\vspace{-8pt}
\caption{Visualization of the denoising effect of CGDenoiser functioning on enhanced frames. Apparently, images after enhancement are seriously degraded by intricate real noise. With denoising support from CGDenoiser, the quality of frames recovers conspicuously. 
}
\label{fig:denoise_demo}
\vspace{-5pt}
\end{figure}

Figure~\ref{fig:heatmap} shows the heatmap of classification confidence from trackers. The intricate noise seriously distracts the attention of trackers in patch (a), making trackers highly sensitive to the temporary occlusion in patch (b) and (c). With the assistance of CGDenoiser, the classification of the foreground becomes much more precise and attentive,  clearly recognizing and coordinating the tracking targets.

\begin{figure}[!t]
\subfloat{\includegraphics[width=0.5\linewidth]{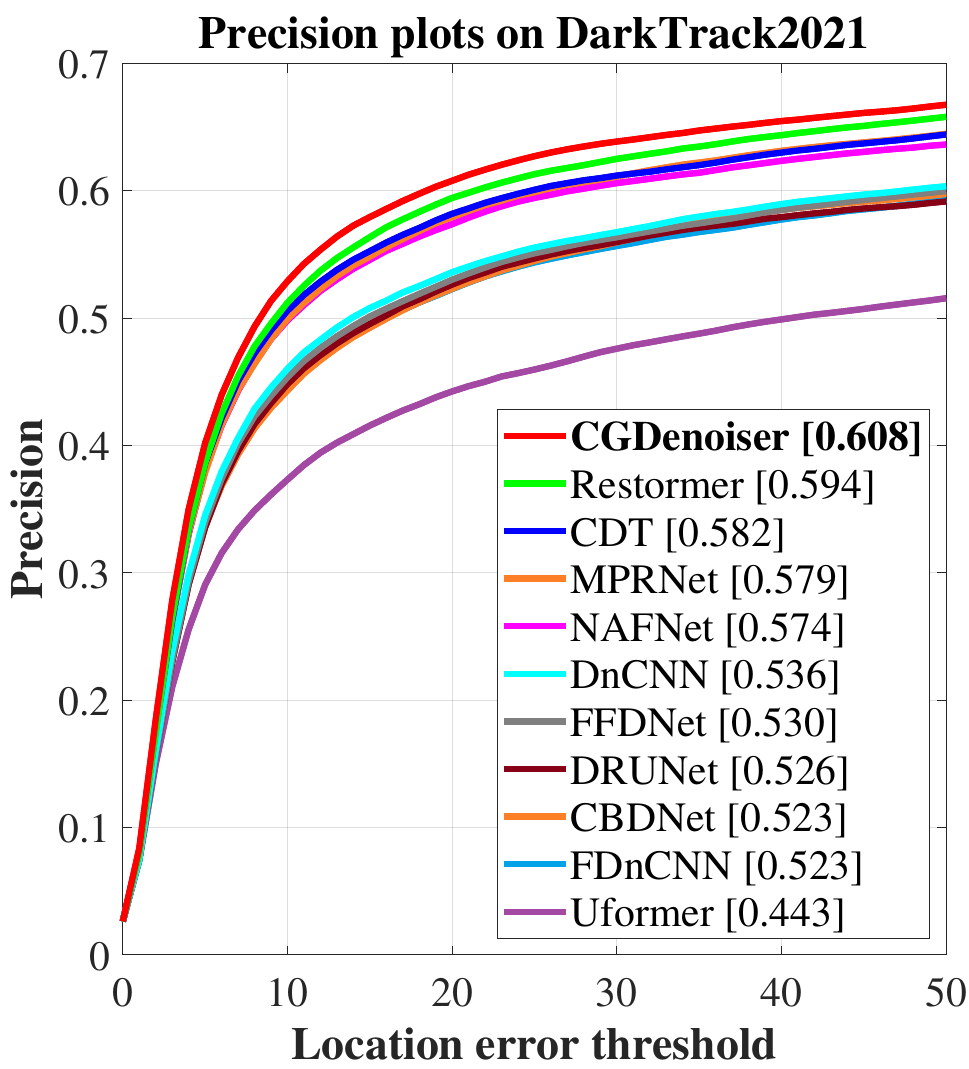}%
}
\subfloat{\includegraphics[width=0.5\linewidth]{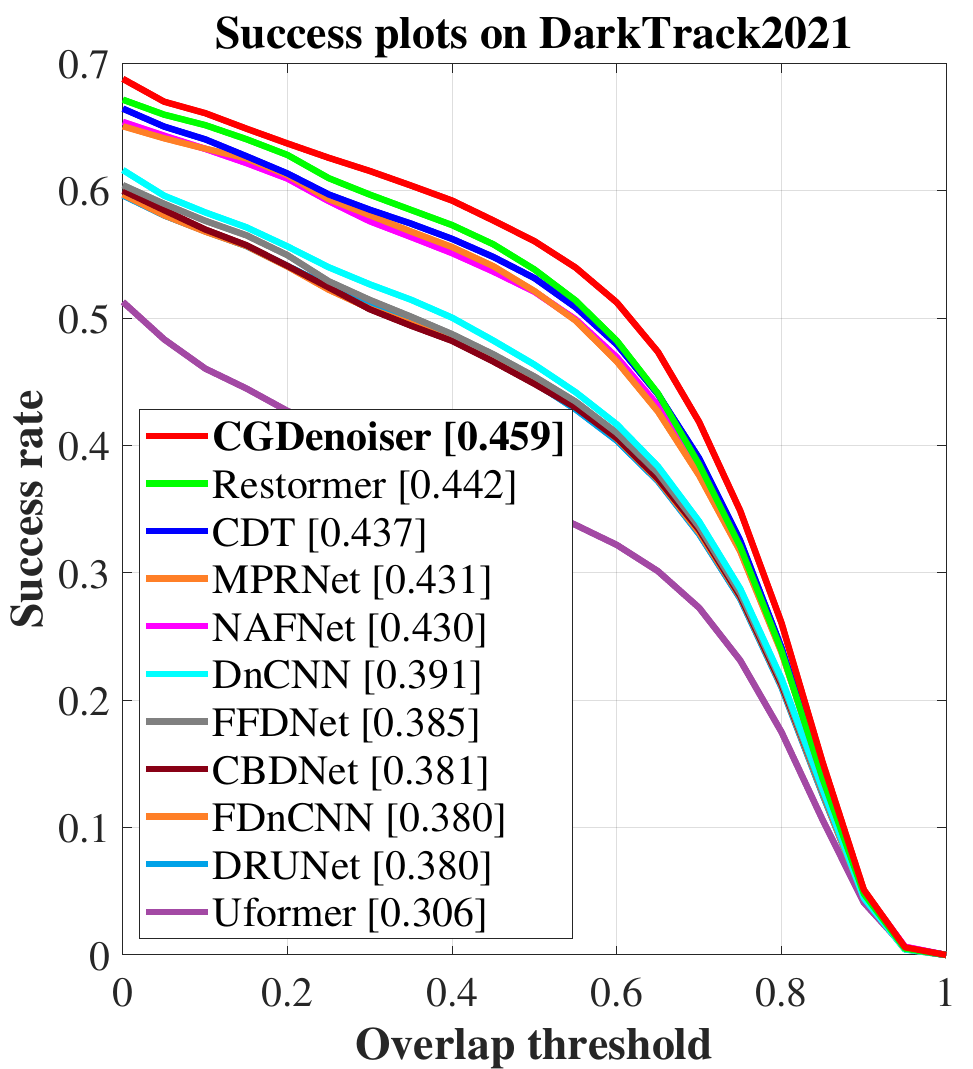}
}
\vspace{-8pt}
\caption{Tracking evaluation on authoritative nighttime benchmark~\cite{Ye2022TrackerMN} of CGDenoiser and multiple leading-edge denoisers~\cite{Zhang2017BeyondAG,Mehri2021MPRNetMP,Wang2022UformerAG,Zamir2022RestormerET,Lu2023CascadedDT,Chen2022SimpleBF,Zhang2018TowardAF}. CGDenoiser obtains the highest performance on both tracking success rate and precision.
}
\label{fig:Tracking performance compare}
\end{figure}

\begin{figure}[t]
\centering
\subfloat{\includegraphics[width=1\linewidth]{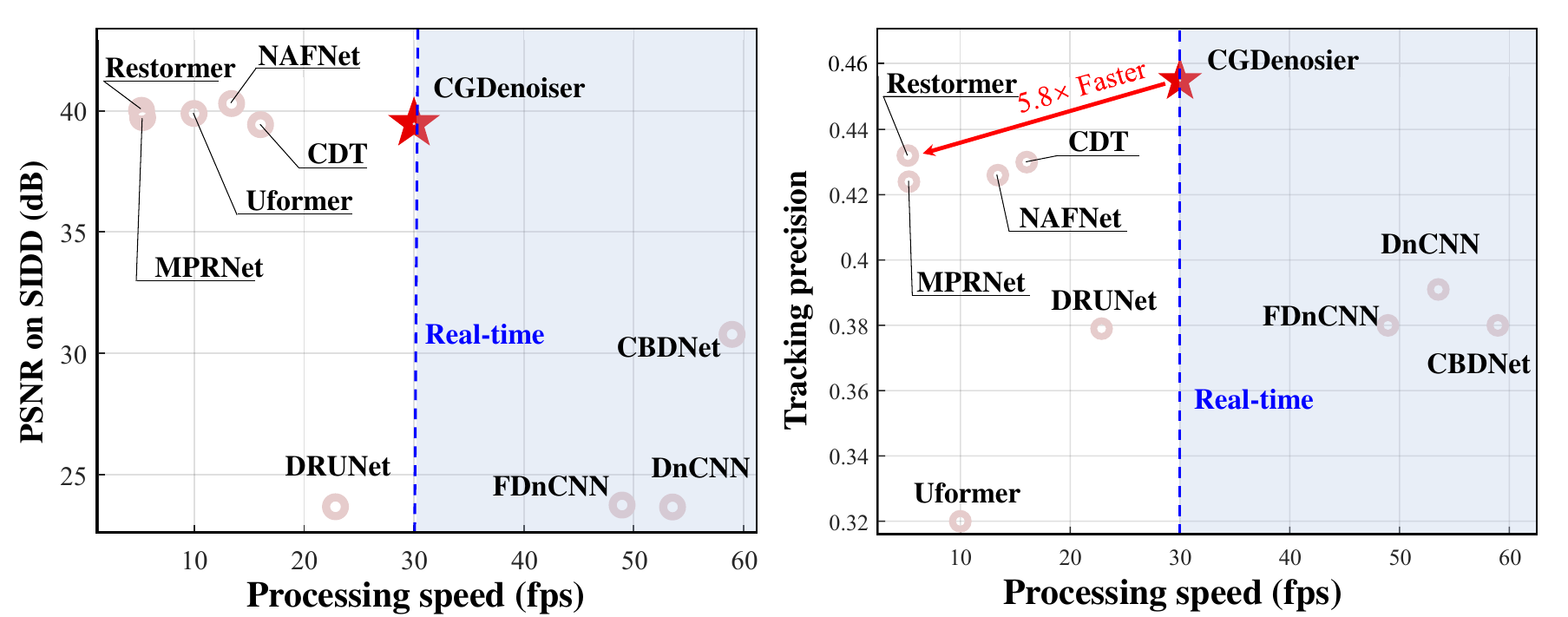}}
\vspace{-8pt}
\caption{The processing speed and the corresponding tracking and denoising performance comparison between CGDenoiser and leading-edge denoisers. 
Compared to the second well-performed denoiser, CGDenoiser works $5.8\times$ faster and supports real-time denoising practice.
}
\vspace{-6pt}
\label{fig:Processing speed}
\end{figure}

\subsection{Comprehensive Comparison with SOTA Denoisers}\label{Comparison with SOTA Denoisers}

\subsubsection{Tracking performance}
To further demonstrate the superiority of the proposed method, comprehensive contrast experiments with multiple SOTA denoisers regarding tracking performance, processing speed, and denoising effect are performed. In the following comparative analysis, the tracking performance is obtained by SiamRPN++~\cite{Li2019SiamRPN++EO} on DarkTrack2021 benchmark~\cite{Ye2022TrackerMN}. Frames are enhanced by DCE++~\cite{Li2022LearningTE} and denoised by different denoisers. The processing speed is tested on 1 NVIDIA RTX3060 GPU and the PSNR values are evaluated on SIDD testset~\cite{Abdelhamed2018AHQ}.

\subsubsection{Denoising performance}
The proposed method aims at better assisting UAV tracking by removing the real noise in nighttime vision, and it yields superior denoising performance. Figure~\ref{fig:denoise_demo} displays several denoising outcomes from the tracking pipeline. The internally intrigued noise is severely amplified by the enhancer and disastrously damages the original frames, which is harmful to tracking feature extraction. Fortunately, the utilization of the proposed method ameliorates this problem to a large extent, effectively restoring images from the complicated noise.
As shown in Fig.~\ref{fig:Processing speed}, the PSNR value on SIDD of CGDenoiser is close to leading-edge denoisers. However, most networks with high denoising performance are impractical for UAV deployment due to low processing speed. CGDenoiser gets rid of over-complicated structures and supports real-time denoising.

The gist of the proposed method lies in promoting nighttime UAV tracking performance. Figure~\ref{fig:Tracking performance compare} has shown the tracking success rate and precision of enhanced SiamRPN++ equipped with various SOTA denoisers, where the proposed CGDenoiser achieves the highest precision gain, demonstrating the excellent capacity of CGDenoiser to eliminate real noise and enormously improve tracking performance.

\subsubsection{Processing speed}
Most current learning-based real-noise denoisers suffer from an enormous amount of parameters and heavy structure, resulting in a considerable decline in processing speed, which makes it impractical to plug denoisers in UAV trackers. 
As shown in Fig.~\ref{fig:Processing speed}, CGDenoiser (\textbf{30 fps}) functions around \textbf{six times} faster than the second well-performed denoiser (5.2 fps) whereas reaching even $\mathbf{5.32\%}$ higher tracking precision, filling the niche for a plug-and-play denoiser with real-time processing speed as well as high tracking performance to deploy in practice.


\subsection{Ablation Study}\label{Ablation Study}

In this section, experiments are extended to prove the effectiveness of different proposed designs. The ablation result can be accessed from TABLE~\ref{tab:ablation}. The Baseline represents the original tracker SiamRPN++ enhanced by DCE++.

\subsubsection*{\bf MKCR}
Unlike traditional changeless post-processing methods, MKCR generates convolutional kernels conditioned on each input, designing signal-based processing strategies for different degradations of different extents. Results in TABLE~\ref{tab:ablation} show that the utilization of MKCR accounts for $5.21\%$ and $14.69\%$ promotion on tracking precision and success rate respectively, which demonstrates the efficacy of the cooperation with MKCR. Additionally, the time cost and model expansion by MKCR are considerably tiny.

\subsubsection*{\bf NRTC}
To better encode the inputs into high dimensional representation as well as promote the processing speed by downsampling initial frames, NRTC is invented to conditionalize the noisy images. 
As shown in TABLE~\ref{tab:ablation}, NRTC results in a $4.50\%$ rise in tracking precision and $14.43\%$ in success rate. The conjunctive gain from the combined implementation of MKCR and NRTC further demonstrates the effectiveness and compatibility of the structure design.

\subsection{Real-World Tests}\label{Real-World Tests}

\begin{figure}[!t]
\centering
\includegraphics[width=1\linewidth]{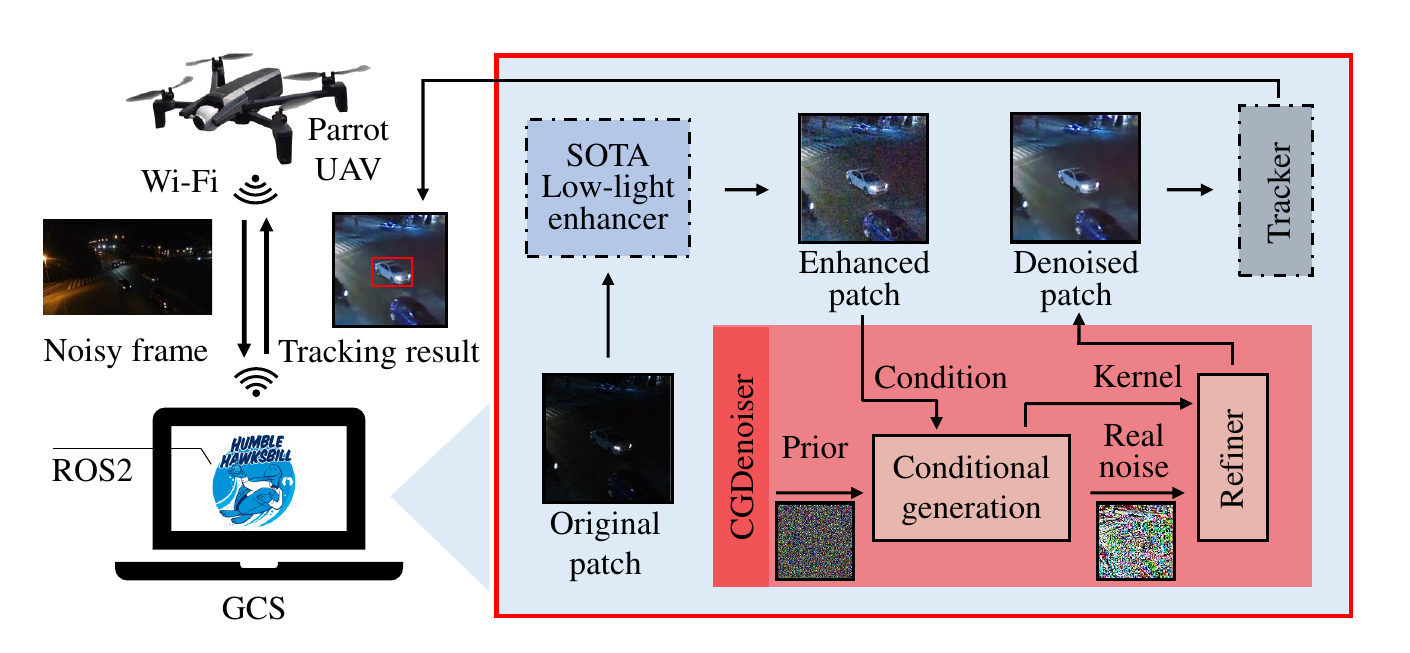}%
\vspace{-5pt}
\caption{
Real-world workflow of nighttime UAV tracking with the aid of CGDenoiser. The ground control station (GCS) receives and processes images from the onboard camera in real time, where the noisy frames are first enhanced and then denoised by the proposed CGDenoiser for the tracker. The tracking results in turn instruct the control of Parrot UAV.
}
\label{fig:real_world}
\end{figure}

\begin{figure}[!h]
\centering
\subfloat{\includegraphics[width=1\linewidth]{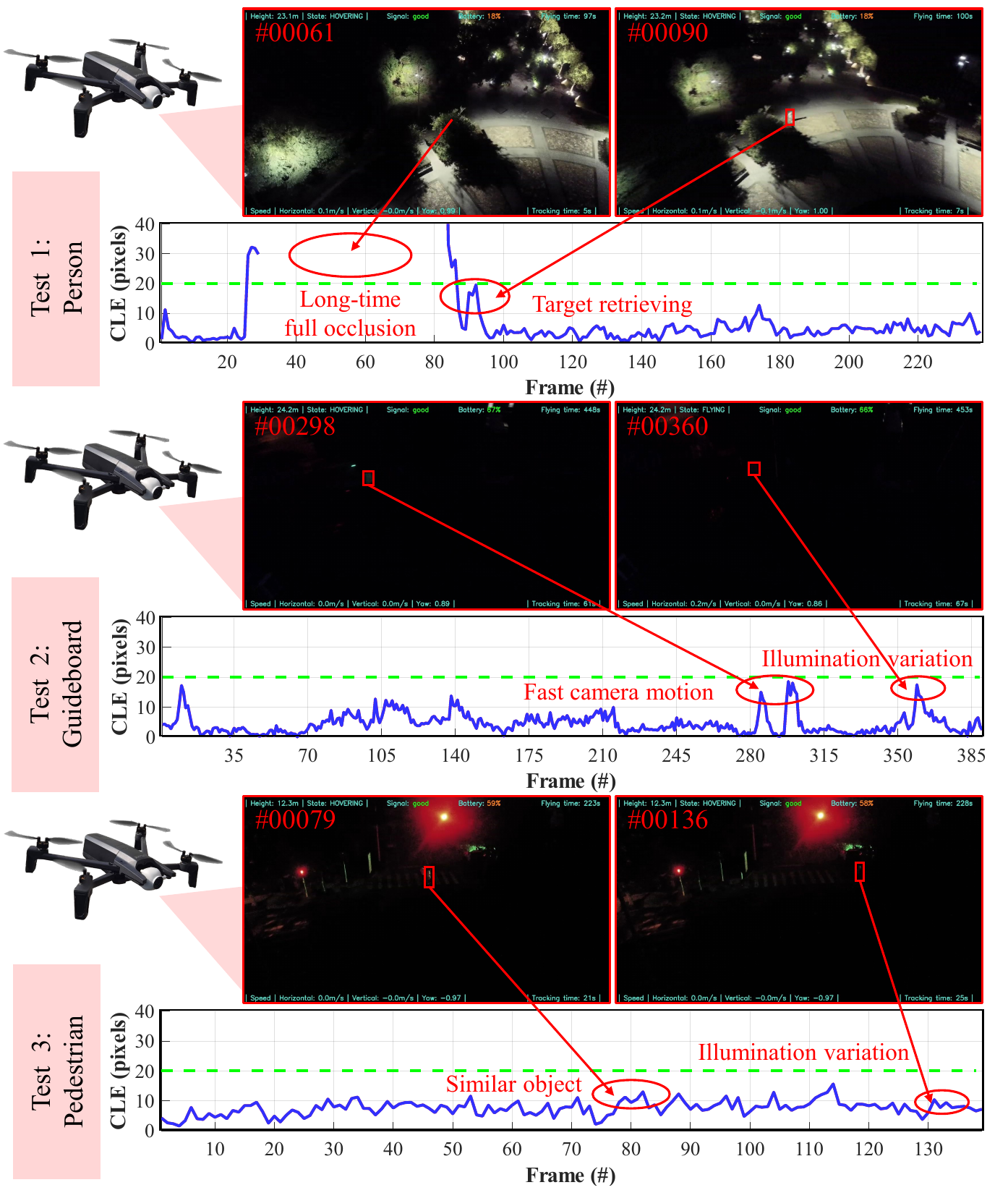}}
\vspace{-10pt}
\caption{Real-world tracking test of challenging sequences.  All results are regarded as successful with CLE below the common threshold (dotted line in \textcolor{green}{green}), validating the outstanding practicality and competence of the proposed method in nighttime UAV tracking.
}
\label{fig:cle}
\vspace{-10pt}
\end{figure}

Real-world tests with a variety of challenges are conducted to prove the practicability of the proposed denoiser. As shown in Fig.~\ref{fig:real_world}, the Parrot UAV\footnote{See \url{https://www.parrot.com/}} is used to acquire sequences in complex nighttime scenes. Frames are transmitted to and processed by a ground control station (GCS) equipped with an NVIDIA RTX3060 GPU in real time, whose processing results in turn assist the pose adjustment of the UAV through WiFi communication. To demonstrate the effectiveness of the proposed method, tracking evaluations of sequences with several knotty challenges including target retrieving from long-time full occlusion, illumination variation, fast camera motion, and similar object are exhibited in Fig.~\ref{fig:cle}.
The CLE curves in \textcolor{blue}{blue} represent the error between estimated bounding boxes and the ground truth, whose threshold (dotted line in \textcolor{green}{green}) is generally adopted as 20 pixels.

\vspace{15pt}

\begin{table}[!b]
\vspace{-20pt}
\renewcommand{\arraystretch}{0.9}
\caption{Tracking performance of CGDenoiser with different designed modules on DarkTrack2021~\cite{Ye2022TrackerMN}.
}
\vspace{-3pt}
\label{tab:ablation}
\centering
\begin{tabular}{ p{1.7cm} p{0.9cm}<{\centering} | p{0.6cm}<{\centering} | p{0.6cm}<{\centering} | p{1.5cm}<{\centering} }
\toprule
 &   Baseline  & & & \textbf{CGDenoiser} \\
\midrule
MKCR        &   -    & \checkmark & - & \checkmark \\
NRTC               &   -   &      -     &      \checkmark     & \checkmark \\
\midrule
Prec.   & 0.422    &  0.444   &    0.441   &    \textbf{0.459}   \\
$\Delta_{p}$ (\%)      &   -   &  5.21    &    4.50    &    \textbf{8.77}    \\
Succ.   & 0.388    &  0.445   &    0.444   &    \textbf{0.455}   \\
$\Delta_{s}$ (\%)      &   -   &  14.69    &   14.43     &    \textbf{17.27} \\
\bottomrule
\end{tabular}
\begin{tablenotes}
\item \textbf{Note:} Prec. and Succ. are respectively the abbreviation of OPE precision (CLE = 20) and success rate, and $\Delta_{p}$ and $\Delta_{s}$ denotes the promotion compared to the Baseline. Data in \textbf{bold} represents results with the best performance.
\end{tablenotes}
\end{table}

\vspace{-3pt}
\section{Conclusions}
\vspace{-4pt}
In this work, a practical plug-and-play denoiser is proposed for better assisting nighttime UAV tracking performance by conditionally generating and removing the real noise in the input frames. Innovatively, the conditional generative method is introduced to alleviate the limitation of supervised training, guiding models to learn the complex distribution of real noise globally. Additionally, MKCR and NRTC are proposed to improve the generation from pre-processing stage and post-processing stage respectively. Comprehensive experiments have demonstrated the high practicality of CGDenoiser, boasting superior tracking performance and fast processing speed.
Real-world tests prove the feasibility of actual nighttime UAV tracking deployment.
To conclude, we strongly believe the practicality of CGDenoiser as a plug-and-play accessory for trackers will conduce to the utilization of vision information in UAV-related intelligent operating systems.

\vspace{15pt}
\section*{Acknowledgment}

This work is supported by the National Natural Science
Foundation of China (No. 62173249) and the Natural Science
Foundation of Shanghai (No. 20ZR1460100).
\vspace{15pt}

\bibliographystyle{IEEEtran}
\bibliography{CGD}

\end{document}